\documentclass[letterpaper]{IEEEtran}
\usepackage[inkscapeformat=png]{svg}
\usepackage{setspace}
\usepackage{extarrows}
\usepackage{graphicx}
\usepackage{amsmath}
\usepackage{amssymb}
\usepackage{booktabs}
\usepackage{epstopdf}
\usepackage{flushend}
\usepackage{colortbl}
\usepackage{array}
\usepackage{multirow}
\usepackage[lined]{algorithm2e}
\SetKwComment{Comment}{//}{}
\usepackage{extarrows}
\usepackage{caption}
\usepackage[caption=false]{subfig}
\usepackage{amsmath}
\usepackage{amssymb}
\usepackage{booktabs}
\usepackage{listings}
\usepackage{svg}
\usepackage{xcolor}
\usepackage{threeparttable}
\usepackage{enumerate}
\usepackage{float}
\usepackage{pgfplots}
\usepackage{bbm}
\pgfplotsset{compat=1.3}
\usetikzlibrary{plotmarks}
\usetikzlibrary{arrows.meta}
\usepgfplotslibrary{patchplots}
\usepackage{grffile}
\usepackage{amsmath}
\usepackage{adjustbox}
\usepackage{framed}
\usetikzlibrary{spy}
\usepgfplotslibrary{groupplots}
\pgfplotsset{width=0.8\linewidth,compat=1.9}
\usepackage{hyperref}
\hypersetup{
    colorlinks=true,
    linkcolor=black,
    urlcolor=black,
}
\usepackage{epstopdf}
\usepackage{gensymb}
\usepackage{textcomp}
\usepackage{amsmath}
\usepackage{bm}
\def\degree{${}^{\circ}$}
\usepackage{cite}
\usepackage{color}

\newcommand{\etal}{\textit{et al}. }
\newcommand{\ie}{\textit{i}.\textit{e}. }
\newcommand{\eg}{\textit{e}.\textit{g}. }

\begin{document}
\captionsetup[figure]{name={Fig.},labelsep=period,singlelinecheck=off} 

\title{
A Spatial Calibration Method for Robust Cooperative Perception
}

\author{Zhiying Song, Tenghui Xie, Hailiang Zhang, Jiaxin Liu,  Fuxi~Wen,~\IEEEmembership{Senior~Member,~IEEE},  Jun Li
\thanks{
Manuscript received: November, 9, 2023; Revised January, 26, 2024; Accepted February, 20, 2024.}
\thanks{This paper was recommended for publication by Editor M. Ani Hsieh upon evaluation of the Associate Editor and Reviewers' comments.
This work was supported by the National Key R$\&$D Program of China under Grant 2021YFB1600402 and 2020YFB1600303.}
\thanks{
The authors are with the School of Vehicle and Mobility, Tsinghua University, Beijing, China. \textit{Corresponding author}: Fuxi Wen, \tt{\footnotesize wenfuxi@tsinghua.edu.cn}.
}
\thanks{Digital Object Identifier (DOI): see top of this page.}
}
\markboth{IEEE Robotics and Automation Letters. Preprint Version. February, 2024}
{Song \MakeLowercase{\textit{et al.}}: A Spatial Calibration Method for Robust Cooperative Perception} 
\maketitle

\begin{abstract}
Cooperative perception is a promising technique for intelligent and connected vehicles through vehicle-to-everything (V2X) cooperation, provided that accurate pose information and relative pose transforms are available. Nevertheless, obtaining precise positioning information often entails high costs associated with navigation systems. {Hence, it is required to calibrate relative pose information for multi-agent cooperative perception.} This paper proposes a simple but effective object association approach named context-based matching ($\tt{CBM}$), which identifies inter-agent object correspondences using intra-agent geometrical context. In detail, this method constructs contexts using the relative position of the detected bounding boxes, followed by local context matching and global consensus maximization. The optimal relative pose transform is estimated based on the matched correspondences, followed by cooperative perception fusion. Extensive experiments are conducted on both the simulated and real-world datasets. Even with larger inter-agent localization errors, high object association precision and decimeter-level relative pose calibration accuracy are achieved among the cooperating agents. 
Demo video, code, and more up-to-date information are available at \href{https://github.com/zhyingS/CBM}{{https://github.com/zhyingS/CBM}}.

\end{abstract}


\section{Introduction}
\label{section:intro}

\IEEEPARstart{C}{ooperative} perception has emerged as a prominent research topic
for intelligent and connected vehicles in recent years \cite{huang2023v2x}.
This technique enhances the perception capability of the individual vehicles by leveraging complementary information from neighboring agents, \eg, vehicles \cite{Wang2020}, drones \cite{liu2020who2com} or infrastructure nodes \cite{khamooshi2023cooperative}. 

\begin{figure}
\centering
\includegraphics[width=0.98\linewidth]{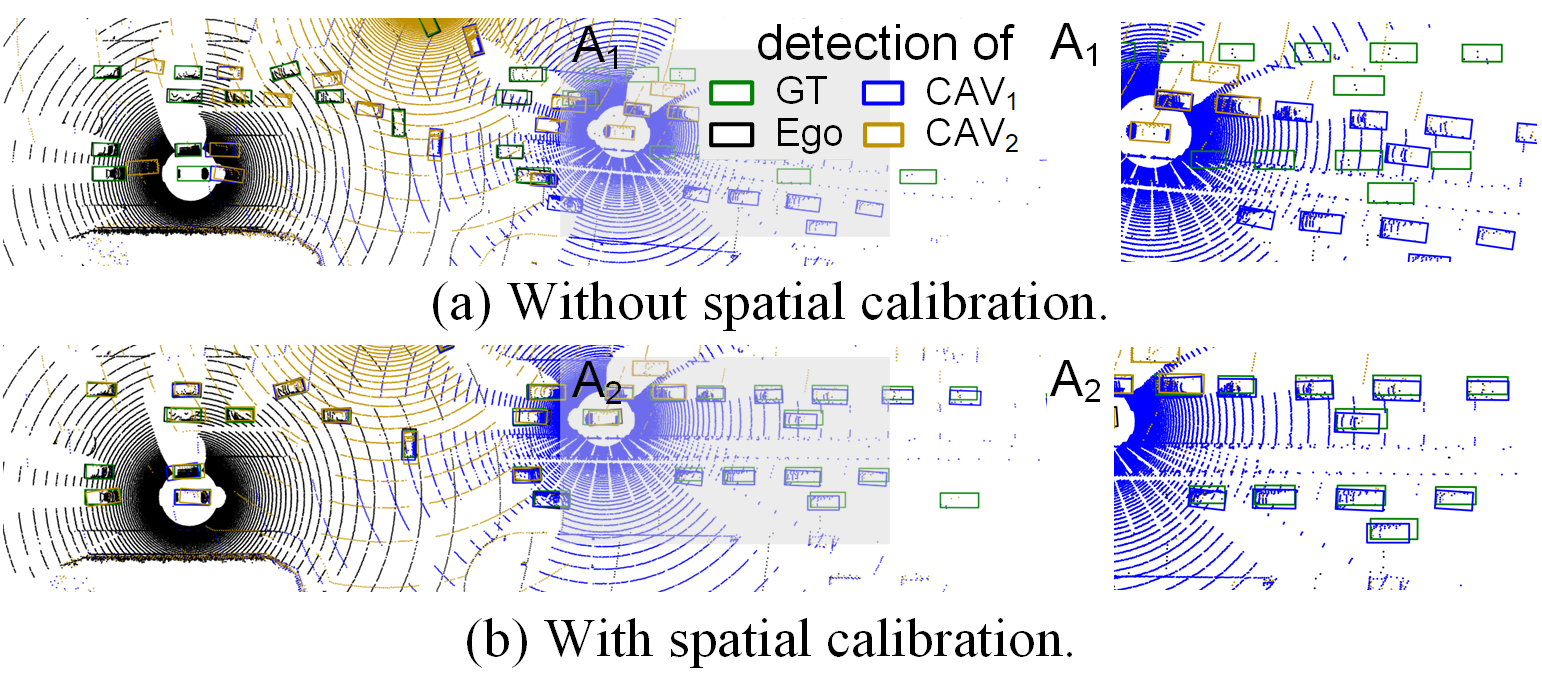}
\caption{Effect of spatial calibration on cooperative perception.
}
\label{fig:system}
\end{figure}

However, aligning perception results among these agents hinges on the availability of precise localization measurements, which can be challenging to obtain in complex traffic environments. Therefore, a spatial calibration module is needed to refine the spatial offset caused by localization errors \cite{caillot2022survey}. 
As an illustration, Fig. \ref{fig:system} shows a cooperative perception scenario of three vehicles. The detection results are transformed into the Ego frame using transformation matrices acquired from localization systems. Errors in this transform result in significant misalignment of the detection results. In such a case, cooperative perception not only fails to improve Ego perception but also disrupt it.

Previous research has suggested calibrating spatial errors through the alignment of raw data \cite{kim2015sensor,Fang2022}, or the use of specific features \cite{vadivelu2021learning,yuan2022leveraging}. Nevertheless, in cooperative perception systems, the preference leans toward a lightweight method that requires minimal information transmission and involves the fewest additional feature extractors.
In this paper, we propose to use only the object-level features, \ie, the detection results in the form of bounding boxes, to achieve a robust calibration of spatial errors. 

Current object-level calibration methods have faced difficulties owing to the following three challenges \cite{Song2023,zhang2022fast,shi2022vips}.  The first challenge is the scarcity of information. Object-level features are extracted by the object detection module, encompassing solely the position, orientation, and dimensions of the surrounding objects. In contrast to using raw data or deep features, this approach results in the loss of a considerable amount of semantic information. The second challenge is the presence of perception errors. The object-level features can be noisy due to the inherent limitations of the detection modules. For example, an object may be detected at an incorrect position with a significantly deviated heading angle.
The perception noise further erodes the usability of object-level features. 
The third challenge lies in the non-co-visible objects. To achieve alignment of inter-agent features, it is crucial to identify the same object from different perspectives. However, a substantial portion of the objects are non-co-visible, meaning they are only visible to one of the cooperating agents. This results in a high proportion of outliers when performing inter-agent object association.

In this paper, we propose a novel inter-agent pose alignment module for object-level distributed cooperative perception systems.
The core idea is to identify inter-agent objects by intra-agent geometrical context.  
The method consists of three steps. 
First, an intra-agent context matrix for each object is constructed by encoding their geometrical features. 
Then, the unique correspondences between inter-agent objects are identified by seeking global consensus among the per-object context matrices. 
Finally, with these global correspondences, the relative transformation matrix is estimated and multi-view perception results are fused into an unified frame, resulting in a spatial error-calibrated cooperative perception output. 

The proposed approach maximizes the utilization of features by embedding information from all objects into each object's local context matrix. 
Each context matrix, constructed from a local perspective, exhibits substantial distinctiveness, enhancing the method's resilience against outliers (non-co-visible objects). 
In addition, the perception errors are efficiently managed by seeking global consensus with redundancy among the context matrices of all the objects.

Our contributions are summarized as follows:
\begin{itemize}
\item We proposed a novel spatial calibration approach for distributed cooperative perception systems. Instead of assuming perfect spatial synchronization and ideal perception, we take localization and perception errors into account and design a system robust to them in complex traffic scenarios.

\item We propose an {effective} inter-agent object association approach that is resilient to perception errors and outliers, achieved by taking into account the distinctive characteristics of objects within transportation scenarios.

\item We achieve decimeter-level spatial calibration using only bounding boxes, rather than raw data or complex features. The proposed method can be naturally extended to V2X-based scenarios, with minimal communication overhead and cost-effective implementation.

\end{itemize}

\section{{Related work}}


\textbf{Cooperative perception.}
Recent research on multi-agent cooperative perception is mainly focused on improving efficiency, performance, robustness, and safety of the process \cite{han2023collaborative,huang2023v2x}. 
Significant progress in improving the detection performance of cooperative perception under ideal cases has been achieved in \cite{Chen2019,Wang2020,Xu2022a,huwhere2comm,yuan2022keypoints,leiyang2023,zhao2023bm2cp}.
For robustness, cooperative perception has been investigated for various issues, such as communication issues \cite{tu2021adversarial,wang2023data,li2023learning}.
In terms of localization errors, V2VNet \cite{Wang2020} was the first to demonstrate the sensitivity of cooperative perception to imperfect localization. Subsequently, many state-of-the-art cooperative perception models, such as those in \cite{xu2022,huwhere2comm,yang2023spatio,zhao2023bm2cp}, have either demonstrated or emphasized this vulnerability. However, efficient solutions in this regard remain elusive.

\textbf{Spatial calibration.}
Researchers tried to calibrate the localization errors using various features.
Vadivelu \etal proposed a learning-based method to encode the sensor data to spatial feature maps and performed pose regression on them \cite{vadivelu2021learning}. 
Yuan \etal selected bounding boxes, points of poles and points of big planar structures as features and developed a RANSAC-based inter-vehicle pose correction method \cite{yuan2022leveraging}.
Yang \etal proposed a feature descriptor for point cloud based on gridded Gaussian distribution with Wasserstein distance for global pose
initialization \cite{Yang2023}. 
TrajMatch calibrates inter-LiDAR pose at the roadside using trajectory and semantic features generated in the object detection/tracking phase \cite{ren2023trajmatch}.
However, these methods require the extraction of specific features for calibration, which represents an additional burden for cooperative perception systems.
Several studies, such as \cite{rauch2013inter, Song2023}, have attempted to use only the results of perception systems and 
tackle inter-vehicle object association using lightweight point cloud registration algorithms. However, these methods often rely on a well-guessed initial pose and are unable to handle larger localization errors, thus limiting their effectiveness in complex scenarios.

\begin{figure}[tb]
\centering
\includegraphics[width=\linewidth]{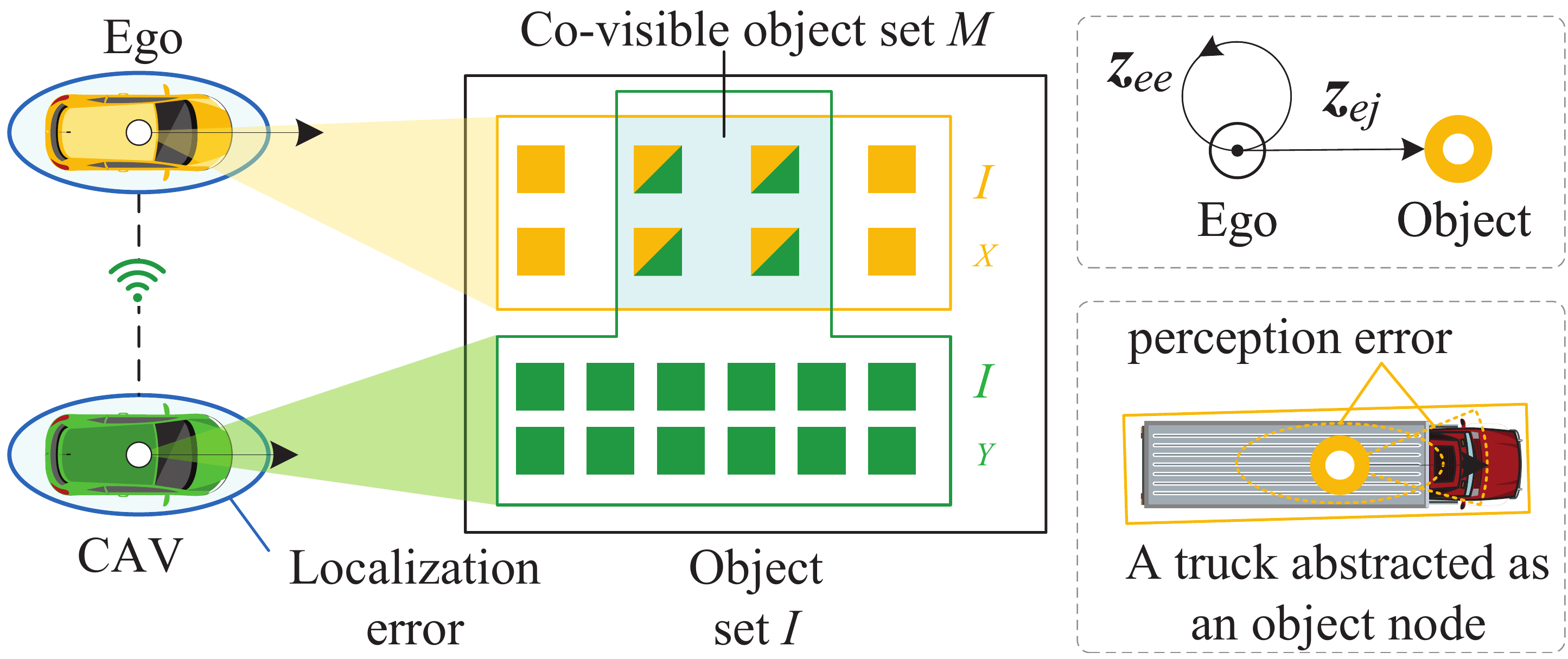}
\caption{{Illustration of the object sets.}}
\label{fig:problem}
\end{figure}

\textbf{Object association.}
Iterative closest point (ICP) and its variants \cite{Besl1992} are commonly used to associate dense object clusters, for example, in \cite{dong2023lidar}. Nevertheless, the objects in cooperative perception are always distributed sparsely.
Recently graph matching-related methods have gained popularity, they rely less on absolute positioning and instead use relative information between nodes for association. 
Gao \etal formulated the association problem in cooperative perception as a non-convex constrained graph optimization problem and developed a sampling-based algorithm to solve it \cite{gao2021regularized}. However, the complexity and time-consuming nature of this approach hinders its applicability. Tedeschini \etal proposed to use a neural network to encode graph node and edge features for association \cite{tedeschini2022addressing}. A computationally efficient method named VIPS is proposed in \cite{shi2022vips} to solve the similar graph optimization problem that makes it available to infrastructure-assisted cooperative perception. %
However, they face challenges in relying on the design of similarity functions and constraint relaxation, handling perception errors and outliers, as well as tuning numerous hyperparameters.

\begin{figure*}[t]
\centering
\includegraphics[width=\linewidth]{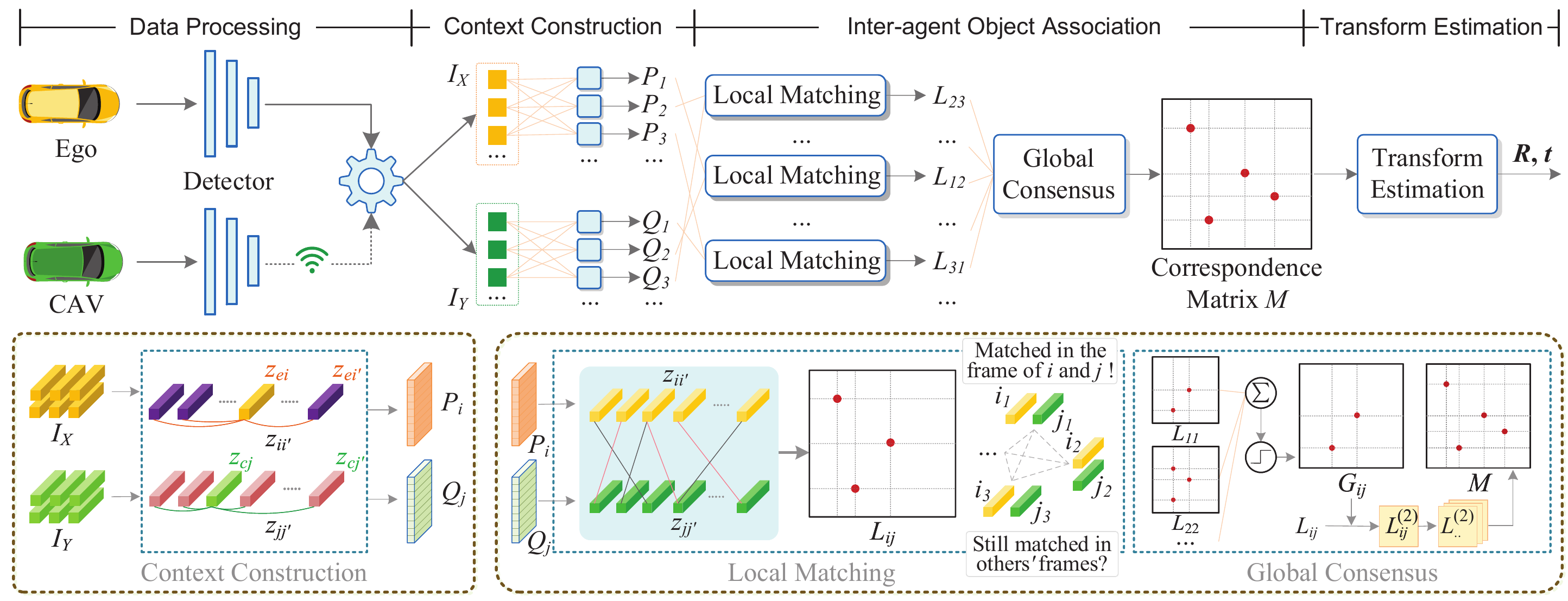}
\caption{
{Framework of the proposed method \tt{CBM}}. Object sets $\mathcal{I}_{\mathcal{X}}$ and $\mathcal{I}_{\mathcal{Y}}$ are detected by the onboard perception system of the Ego vehicle and CAV, respectively. Subsequently, the Ego establishes context based on $\mathcal{I}_{\mathcal{X}}$ and $\mathcal{I}_{\mathcal{Y}}$. Preliminary correspondences between objects in $\mathcal{I}_{\mathcal{X}}$ and $\mathcal{I}_{\mathcal{Y}}$ are identified via the local matching module, and subsequently refined through the global consensus module. Finally, the relative pose between the Ego and CAV is estimated.
}
\label{fig:framework}
\end{figure*}

The rest of the paper is organized as follows: In Section \ref{sec:method}, the spatial calibration approach is proposed. Section \ref{sec:sind} and \ref{sec:opv2v} present the experimental evaluation of the real-world dataset SIND and simulated cooperative perception dataset OPV2V, respectively. 
Finally, Section \ref{section:conclusion} summarizes the conclusions. {The framework of the proposed method is shown in Fig. \ref{fig:framework}.}

\section{method}
\label{sec:method}

\subsection{Problem formulation}
\label{sec:pro}

As illustrated in Fig. \ref{fig:problem}, consider a vehicular network consisting of two cooperative nodes $\mathcal{A}=\left\{e,c\right\}$ (one Ego agent and one cooperative agent, \eg, connected vehicles or RSU) and some passive nodes denoted as {objects $\mathcal{I}$ with cardinality $N$, which can be further segmented into two overlapped groups: objects  $\mathcal{I}_{\mathcal{X}}$ with cardinality $N_e$ that can be sensed by Ego agent and objects $\mathcal{I}_{\mathcal{Y}}$ with cardinality $N_c$ that are accessible for sensing by the cooperative agent. In the presence of co-visible objects, the overlap between $\mathcal{I}_{\mathcal{X}}$ and $\mathcal{I}_{\mathcal{Y}}$ represents the co-visible object set $\mathcal{M}$.
}

{The state of a node (either agent node or object node) defined in the world coordinate system is denoted by vector $\bm{s}=[\bm{p}^T,\bm{d}^T,\theta]^T\in \mathbb{R}^6$, which includes Bird's-eye-view position $\bm{p}\in \mathbb{R}^2$,  orientation ${\theta}\in \left( 0,2\pi\right)$ and 3D dimension size (height, width and length) $\bm{d}\in \mathbb{R}^3$. 
We denote the measurement vector as 
$$
\bm{z}_{kh}=\left[\left(\bm{z}_{kh}^{\bm{p}}\right)^T,\left(\bm{z}_{kh}^{\bm{d}}\right)^T,\bm{z}_{kh}^{\theta} \right]^T \in \mathbb{R}^6, \forall \  k \in \mathcal{A}, h \in \mathcal{A} \cup \mathcal{I}
$$
The notations $\bm{z}_{kh}$ for $k \neq h$ represent inter-node (relative) measurements, while $\bm{z}_{kk}$ represents the intra-node (absolute) measurements of agent $k \in \mathcal{A}$, as shown in Fig. \ref{fig:problem}.
}

{\textbf{Intra-node Measurements. }}The position and orientation states of agents are measured by their onboard localization and navigation systems, by
\begin{equation}
{\bm{z}_k \triangleq} \bm{z}_{kk}= \bm{g}^{(k)}(\bm{s}_k)+\bm{\omega}_k, \ \forall \  k\in \mathcal{A}
\label{eq:intra_measure}
\end{equation}
where $\bm{g}^{(k)}(\cdot)$ denotes a function of agent absolute localization states, and $\bm{\omega}_k$ represents the localization noise.

{\textbf{Inter-node Measurements. }}The states of the objects are measured by the perception systems of the agents, by  
\begin{equation}
\bm{z}_{kh}=\boldsymbol{h}^{(k)}\left(\boldsymbol{s}_k,\boldsymbol{s}_h\right)+\bm{\omega}_{kh}, \; {\forall \  k \in \mathcal{A}, h \in \mathcal{I}}
\label{eq:measurement}
\end{equation}
{where $\boldsymbol{h}^{(k)}(\cdot)$ denotes the perception system}, and $\bm{\omega}_{kh}$ represents the  perception noise.  

Given the above measurements, \ie, the state measurement of both agents $\bm{z}_k$ and the detection measurement of objects $\bm{z}_{kh}$, the problem is to estimate the transformation matrix $\bm{T}_c^e$ between the coordinate of the Ego agent and the cooperative agent. 
We frame this problem as a measurement alignment issue, which is then subdivided into two components: inter-agent object association and transform estimation. The former is to find the matching set of co-visible objects. {From a ground truth perspective, the object set $\mathcal{I}$ can be categorized into two groups: co-visible objects $\mathcal{M}=\mathcal{I}_{\mathcal{X}}\cap\mathcal{I}_{\mathcal{Y}}$ that can be jointly detected by both agents and non-co-visible objects  $\overline{\mathcal{M}}=\mathcal{I}-\mathcal{M}$ that can only be sensed by one of the agents.} However, these two sets are not directly observable from agents' local measurements. The inter-agent object association task is to estimate the co-visible object set $\hat{\mathcal{M}}$ from the local measurement sets, followed by the second task, \ie, inter-agent transform estimation,
\begin{align}
{
\hat{\bm{R}_c^e},\hat{\bm{t}_c^e} = \arg \min ||\left(\bm{R} \cdot \bm{z}_{ch}^{\bm{p}} + \bm{t}\right) -\bm{z}_{eh}^{\bm{p}}||_2, \ h\in \hat{\mathcal{M}}
}
\end{align}
{where $\hat{\bm{R}}_c^e\in SO(2)$ is a rotation matrix  and $\hat{\bm{t}}_c^e\in \mathbb{R}^2$ denotes a translation vector.}%

\subsection{Context-based inter-agent object association}
\label{sec:context}

Corresponding to the index set $\mathcal{I}_\mathcal{X}$ and $\mathcal{I}_{\mathcal{Y}}$ defined in section \ref{sec:pro}, let the state of the object $i \in \mathcal{I}_{\mathcal{X}}$ be $\bm{s}_i = [\bm{p}_i^T,\bm{d}_i^T,\theta_i]^T \in \mathbb{R}^3$, where $\bm{p}_i\in \mathbb{R}^2$ and $\theta_i \in (0,2\pi)$ is the 2D position and
orientation in the Bird’s-eye-view, respectively. $\bm{d}_i\in \mathbb{R}^3$ denotes 3D dimensions. Similarly, for object $j\in \mathcal{I}_{\mathcal{Y}}$, $\bm{s}_j=[\bm{p}_j^T,\bm{d}_j^T,\theta_j]^T$.

The goal of this subsection is to find the covisible object set {$\hat{\mathcal{M}}$ that contains the same objects observed from different agents' view.}
A coarse-to-fine strategy is employed to approximate this matching correspondence. This involves initially identifying coarse matching sets that {include possible results}, followed by the removal of outliers and the attainment of a global consensus, ultimately yielding the final estimation. The details are shown in the following pages.

\subsubsection{Intra-agent context construction}
\label{sec:intra-agent context}

In real-world traffic environments, each traffic participant possesses distinct attributes, such as position, direction, and appearance. Consequently, when viewed from the perspective of an individual vehicle, the surrounding environment is inherently unique, thereby constituting its context. In other words, context uniquely encodes the relationships between nearby objects from an object's local perspective. A simple case is shown in Fig. \ref{fig:context_}. 
Inspired by such an observation, we employ context-based comparisons to identify and locate identical objects across multiple views.

Similar to the preprocessing procedure in \cite{Song2023}, we first standardize the measurements by converting them into the Ego frame using transform
\begin{equation}
\widetilde{\bm{T}}_c^e=f\left( \bm{z}_c,\bm{z}_e \right)
\end{equation}
{for $c,e \in \mathcal{A}$}, where $\bm{z}_c$ and $\bm{z}_e$ are defined in  (\ref{eq:intra_measure}), the transform function $f(\cdot)$ can be found in \cite{Song2023}, representing the transform calculated from on-board localization systems. 
$\widetilde{\bm{T}}_c^e$ is the desired ${\bm{T}}_c^e$ if there is no localization noise in the measurement of intra-agent position and orientation.
As a result,
\begin{equation}
\bm{z}_c^{(e)}=\widetilde{\bm{T}}_c^e(\bm{z}_{c})
,\ \bm{z}_{cj}^{(e)}=\widetilde{\bm{T}}_c^e(\bm{z}_{cj}),\ \forall j \in \mathcal{I}_{\mathcal{Y}}
\label{eq:transf}
\end{equation}
where the superscript $\cdot ^{(e)}$ indicates that it's in the coordinate of the Ego agent. For brevity in the following text, we will omit the superscript, but it should be understood that all measurement values related to the cooperative agent have been transformed into the Ego coordinate system according to (\ref{eq:transf}).

In the Ego frame, the directions of objects (both in $\mathcal{I}_{\mathcal{X}}$ and $\mathcal{I}_{\mathcal{Y
}}$) are adopted by defining their heading towards the front in the Ego frame as the forward direction.
Then the relative positional measurements between objects in the local frame are given by
{
\begin{equation}
\begin{aligned}
\boldsymbol{z}_{ii'}^{\bm{p}} &= \boldsymbol{R}^T (\bm{z}^{\theta}_{ei})\left (\boldsymbol{z}_{ei'}^{\bm{p}}-\boldsymbol{z}_{ei}^{\bm{p}}\right),\ &\forall i,i'\in\mathcal{I}_\mathcal{X}\\
\boldsymbol{z}_{jj'}^{\bm{p}} &= \boldsymbol{R}^T (\bm{z}^{\theta}_{cj})\left (\boldsymbol{z}_{cj'}^{\bm{p}}-\boldsymbol{z}_{cj}^{\bm{p}}\right),\ &\forall j,j'\in\mathcal{I}_\mathcal{Y}
\end{aligned}
\end{equation}
where $\bm{R}(\theta)\in SO(2)$ is a rotation matrix of a rotation angle $\theta$.
{Note that $\bm{z}_{ii'}^{\bm{p}}$ and $\bm{z}_{jj'}^{\bm{p}}$ contains perception errors in (\ref{eq:measurement}). }

In real-world traffic scenarios, each traffic participant occupies a significant space considering their dimensions and the requirement of maintaining a safe distance between road users. Therefore, even with some measurement errors, their position vectors remain highly distinctive within their occupied space, \ie, the discrimination of $\bm{z}_{ii'}^{\bm{p}}$ ($\bm{z}_{jj'}^{\bm{p}}$) as a feature vector can still be maintained, thereby we define it one of $i$'s ($j$'s) context vectors.

Incorporating all the objects in the vicinity, the context matrices are obtained,
{
\begin{equation}
\begin{aligned}
\bm{P}_i&=\left[\bm z_{i1}^{\bm{p}},\bm z_{i2}^{\bm{p}}, \ldots, \bm z_{i i'}^{\bm{p}}, \ldots\right] &\in \mathbb{R}^{2\times N_e}
\\
\bm{Q}_j&=\left[\bm z_{j1}^{\bm{p}},\bm z_{j2}^{\bm{p}}, \ldots, \bm z_{j j'}^{\bm{p}}, \ldots\right] &\in \mathbb{R}^{2\times N_c}
\end{aligned}
\label{eq:Pi}
\end{equation}}

As context captures the relationships among objects and their neighboring objects, it inherently includes robust spatial constraints between connected objects and remains invariant to rigid transformations. It is worth noting that the concept of context shares similarities with graph descriptors that encode intra-node and inter-node information. However, the context incorporates strong spatial constraints between connected objects, making it more suitable for real-world driving applications where spatial relationships play a crucial role.

\begin{figure}[t]
\centering
\includegraphics[width=\linewidth]{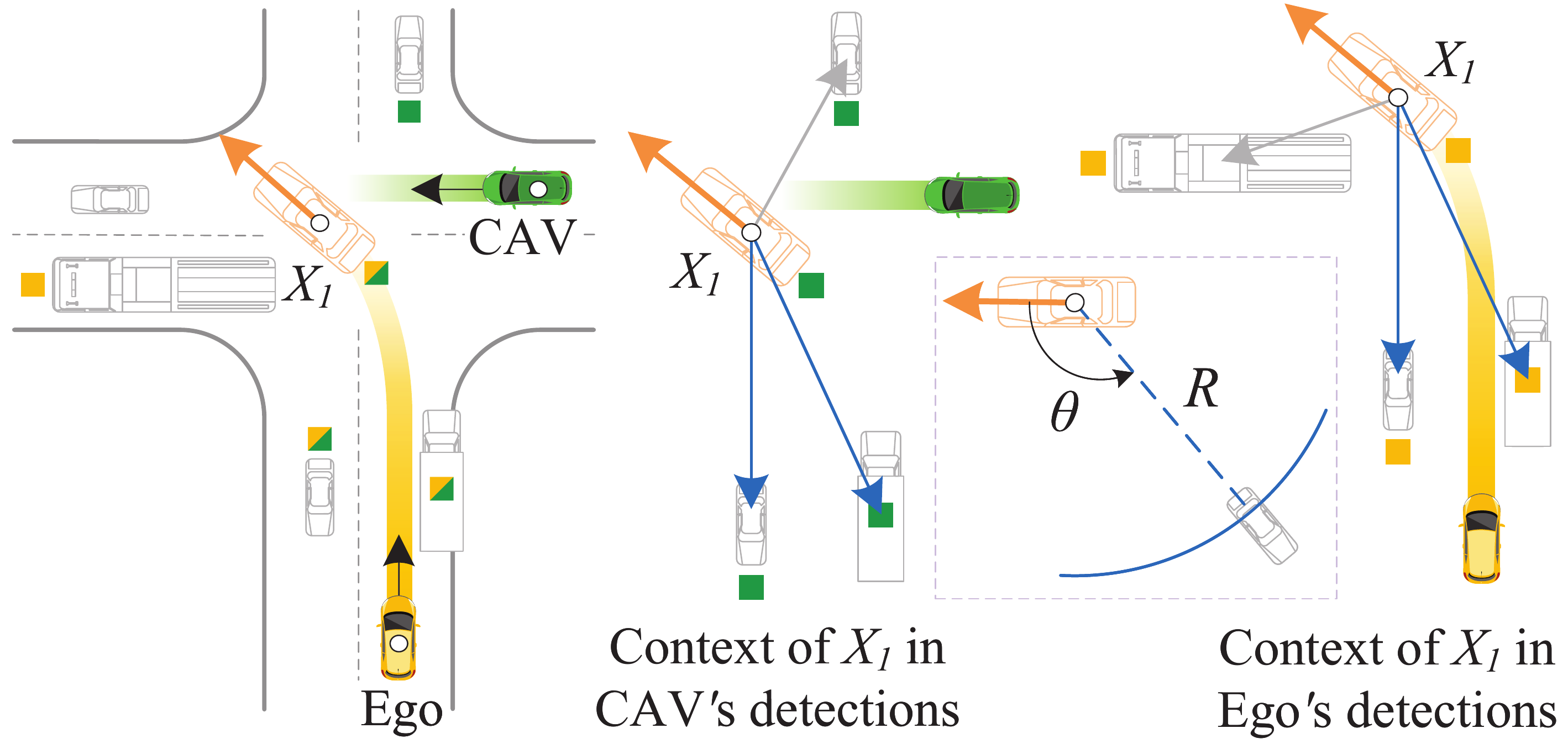}
\caption{
{Context of object $X_1$ in the detections of the Ego and CAV, respectively.
}}
\label{fig:context_}
\end{figure}

\subsubsection{Context similarity-based coarse matching}
\label{section:similarity}

Given $\bm{P}_i$ and $\bm{Q}_j$, the similarity between $\bm{z}^{\bm{p}}_{i i'}$  and $\bm{z}_{j  j'}^{\bm{p}}$ (denoted as $\bm{z}_{i'}$ and $\bm{z}_{j'}$ below) is defined as
\begin{equation}
\label{eq:similarity}
\begin{aligned}
&{S}_{\bm{z}}=\frac{\alpha}{\sigma_1} \left ( \arccos \frac{\left|\boldsymbol{z}_{i'}^T \boldsymbol{z}_{j'}\right|}{\|\boldsymbol{z}_{i'}\|_2\cdot \|\boldsymbol{z}_{j'}\|_2}\right)+\frac{\beta}{\sigma_2} {   \|\boldsymbol{z}_{i'}-\boldsymbol{z}_{j'}\|_1}
\end{aligned}
\end{equation}
where ${S}_{\bm{z}}\in \mathbb{R}$, $||\cdot||_1$ denotes $l_1$ norm. The first term denotes the angular distance and the second characterizes the length difference between the local context of $i$th object in $\mathcal{I}_\mathcal{X}$ and $j$th object in $\mathcal{I}_\mathcal{Y}$. $\alpha>0$ and $\beta>0$ are the parameters to tune the weights of angular and length distance. 
$\sigma_1$ is set to tolerate angular perception errors caused by the positional error of the surrounding objects ($i'$ and $j'$) and the heading angle error of the center object ($i$ and $j$). $\sigma_2$ is set to handle the vector length noise caused by the positional error of both the center and surrounding objects. The use of absolute value operation in the $\left|\boldsymbol{z}_{i'}^T \boldsymbol{z}_{j'}\right|$ term is intended to avoid ambiguity caused by heading direction since detecting the direction of a road user frequently results in opposite judgments.

For the sake of efficiency, we set $\alpha=1$ and $\beta=0$ first to get a preliminary similarity ${S}^{(1)}_{\bm{z}}$, then pick out those highly similar pairs to further compare Euclidean similarity ${S}^{(2)}_{\bm{z}}$. The whole procedure is defined as follows:

\vspace{-4mm}
\begin{algorithm}[!h]
\For{$i\in \mathcal{I}_{\mathcal{X}}$ and $j\in \mathcal{I}_{\mathcal{Y}}$}
{
Initialize an void correspondence set ${\mathcal{L}}_{ij}$\;
\For{$i'\in \mathcal{I}_{\mathcal{X}}$ and $j'\in \mathcal{I}_{\mathcal{Y}}$}
{
\If{${S}_{\bm{z}}^{(1)}={S}_{\bm{z}}(\alpha=1,\beta=0)\leqslant 1$ and ${S}_{\bm{z}}^{(2)}={S}_{\bm{z}}(\alpha=0,\beta=1)\leqslant 1$}{${\mathcal{L}}_{ij}\leftarrow {\mathcal{L}}_{ij} \cup (i',j')$\;}
}}
\end{algorithm}
\vspace{-5mm}

{After these steps, a }preliminary correspondence  ${\mathcal{L}}_{ij}$ for each pair $i\in \mathcal{I}_{\mathcal{X}}$ and $j\in \mathcal{I}_{\mathcal{Y}}$ is obtained, resulting for a local correspondence matrix related to ${\mathcal{L}}_{ij}$ as
\begin{equation}
\bm{L}_{ij}(i',j')=\left\{
\begin{aligned}
&1, \ \text{if } (i',j')\in {\mathcal{L}}_{ij} \text{ and } \text{card}({\mathcal{L}}_{ij})\geq 2. \\
&0, \ \text{otherwise}.
\end{aligned} \right.
\end{equation}
where $\bm{L}_{ij}\in \{0,1 \}^{N_e \times N_c}$, and operator $\text{card}(\cdot)$ denotes counting the number of elements in the set.

By changing the soft thresholds $\sigma_1$ and $\sigma_2$, ${\mathcal{L}}_{ij}$ can encompass a large number of potential matches, aiming to include a significant portion of the ground truth correspondences. The solution of the object association problem can be obtained by filtering outliers from $\bm{L}_{ij}$, and we achieve this by maximizing the global consensus across all $i\in \mathcal{I}_{\mathcal{X}}$ and $j\in \mathcal{I}_{\mathcal{Y}}$.

\begin{figure*}[t]
\centering
\includegraphics[width=0.99\linewidth]{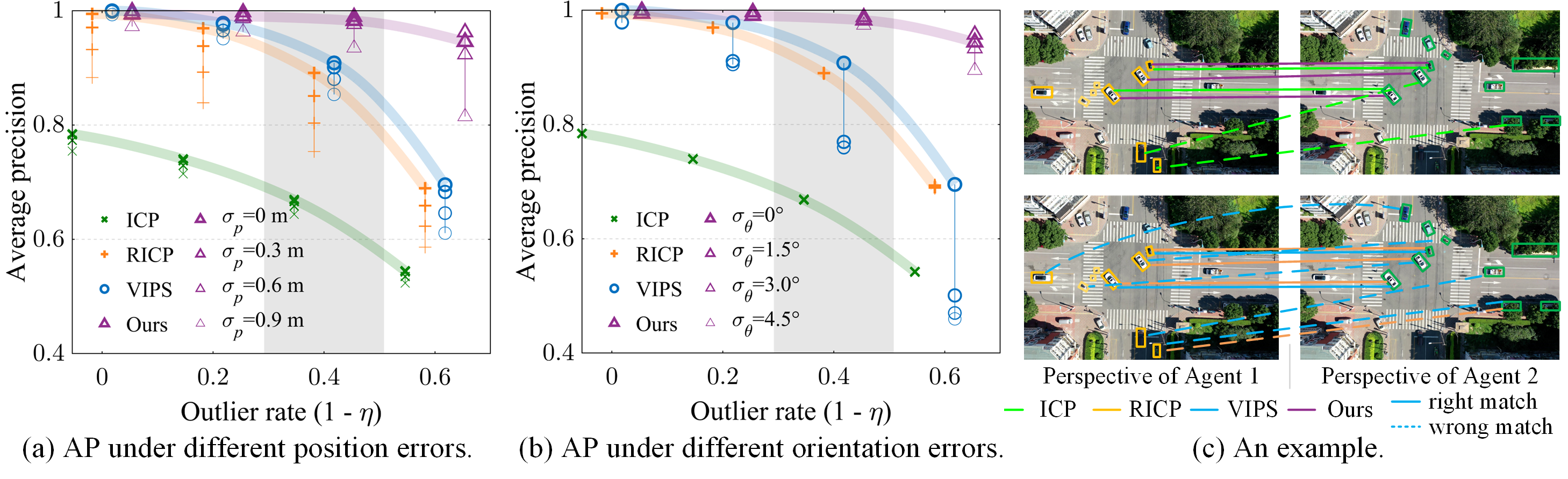}
\caption{{Quantitative results and qualitative demonstration on SIND. 
}}
\label{fig:ap_matching}
\end{figure*}

\subsubsection{Global consensus maximization}
\label{sec:global}

To filter out mismatched correspondences in $\bm{L}_{ij}$, a global filter matrix $\bm{G}_{ij} \in \{0,1\}^{N_e \times N_c}$ is developed for each object pair $i\in \mathcal{I}_{\mathcal{X}}$ and $j\in \mathcal{I}_{\mathcal{Y}}$. The basic idea is to assess the already matched pairs from a global perspective. We eliminate pairs
that are accepted as matched ones in some objects' local frames but not embraced by all the objects, 
\begin{equation}
\bm{G}_{ij}(i',j')=\left\{
\begin{aligned}
1 , \ & \text{ if } \sum_{(k',h')\in \mathcal{L}_{ij}}\bm{L}_{k'h'}(i',j') > 1.\\
0, \ & \text{otherwise.}
\end{aligned}
\right.
\end{equation}
This operation assesses the non-zero correspondences in $\bm{L}_{ij}$ from each other’s perspective to maximize global consensus.
Then the improved correspondence matrix becomes
\begin{equation}
\bm{L}_{ij}^{(1)}=\bm{G}_{ij} \circ \bm{L}_{ij},
\end{equation}
where $ \bm{L}_{ij}^{(1)} \in \{0,1 \}^{N_e \times N_c}$.

To further eliminate one-to-many matching correspondences, where one object in $\mathcal{I}_\mathcal{X}$ matches with several objects in $\mathcal{I}_\mathcal{Y}$, or vice versa, an extra rule is added, then
\begin{equation}
\begin{aligned}
\bm{L}^{(2)}_{ij}(i',j')=\left\{\begin{aligned}
&0,\text{if } \left\{\begin{aligned} &\sum_{j'=1}^{N_c}\bm{L}^{(1)}_{ij}(i',j')>1, \text{ or } \\
&\sum_{i'=1}^{N_e} \bm{L}^{(1)}_{ij}(i',j')>1.\\
\end{aligned}\right. \\
&\bm{L}^{(1)}_{i,j}(i',j'), \text{otherwise. }
\end{aligned} \right.
\end{aligned}
\end{equation}

Finally, the suboptimal matching correspondences can be obtained by 
\begin{equation}
\label{eq:solution_A}
\bm{M}=\arg \max_{i,j} \left\| \bm{L}^{(2)}_{ij}\right\|_0
\end{equation}
where $||\cdot||_0$ is $l_0$ norm. The corresponding matched set is 
\begin{equation}
\hat{\mathcal{M}}=\{(i,j)|i\in \mathcal{I}_{\mathcal{X}},j\in \mathcal{I}_{\mathcal{Y}},\bm{M}(i,j)\neq 0\}.
\end{equation}
{where $i$ and $j$ are the same objects in the real world.}

\subsection{Transform estimation and perception fusion}
\label{sec:estimation}

Given the set of matched object pairs $\forall \ (i,j)\in \hat{\mathcal{M}}$,
if we denote the rotation matrix $\boldsymbol{R}$ and translation vector $\boldsymbol{t}$, then
{
\begin{equation}
\label{eq:proscrustes}
{\boldsymbol{R}}^*, {\boldsymbol{t}}^* =\arg \min_{\boldsymbol{R},\boldsymbol{t}} \ {\sum_{i=1}^{|\hat{\mathcal{M}}|} \psi \left(\left\|\boldsymbol{z}_{ei}^{\bm{p}}- (\boldsymbol{R}\cdot \boldsymbol{z}_{ci}^{\bm{p}}+\boldsymbol{t})\right\|_2\right)}
\end{equation}
where $|\cdot|$ denotes the number of elements in the set, and $\bm{R}\in SO(2)$, $t\in \mathbb{R}^2$.
We adopt the strategy in \cite{zhang2022fast} to design $\psi(x)$ and solve $\bm{R}$ and $\bm{t}$}, please check \cite{zhang2022fast} for the details of the solution. 

Finally, the estimated inter-agent transform matrix is
\begin{equation}
\hat{\bm{T}}_c^e = \begin{bmatrix}
\bm{R}^* & \bm{t}^*\\
0 &1
\end{bmatrix} \cdot \widetilde{\bm{T}}_c^e
\end{equation}

After imposing the calibration transform on the objects detected by the cooperative agent, the objects from multiple views are aligned under the Ego frame { and fused using
Non-maximum suppression (NMS) \cite{Xu2022a, Bodla2017}, which is typically integrated as the final step of the object detection algorithm. }
{Please refer to \cite{Bodla2017} for details.}

\section{Validation on real-world dataset}
\label{sec:sind}

\subsection{Experiments setting}

\textbf{Dataset.}
Due to the nascent stage of cooperative perception technology, most datasets are focused on evaluating object detection performance, and very few datasets are available for evaluating spatial robustness and object association performance. 
We opt to use SIND \cite{xu2022drone}, which is a real-world drone dataset captured from a signalized intersection from a stationary aerial perspective for about 420 minutes. The dataset includes more than 13,000 traffic participants {in various types like cars, pedestrians, and motorcycles.}

\textbf{Metrics and Benchmarks.}
Given the estimated association set $\hat{\mathcal{M}}$ and the ground truth matching set $\mathcal{M}$,  we evaluate the average precision.
Three benchmarks are considered, including Iterative Closest Point (ICP) \cite{Besl1992}, Robust Iterative Closest Point (RICP) \cite{zhang2022fast}, and VIPS \cite{shi2022vips}.
ICP is a fundamental technique for point association. As a classical method,  many variants have occurred recently, among which RICP is the latest achievement. 
VIPS is the state-of-the-art method for inter-vehicle object association using graph matching techniques. Compared with other graph matching-based algorithms, faster processing speed and higher accuracy are achieved for VIPS. 

\textbf{Co-visible objects. }
In real-world traffic scenarios, non-covisible objects exist due to a limited field of view and occlusions. These objects are outliers that have a serious impact on matching tasks.
Given the object index set $\mathcal{I}$ at a single frame, 
we randomly sampled the co-visible object set $\mathcal{M}$ to simulate cooperative perception, such that
$\text{card}\left(\mathcal{M}\right)=\eta \cdot \text{card}\left(\mathcal{I}\right) \ ,  \mathcal{M} \subseteq \mathcal{I}
$, 
where $\eta$ is the rate of co-visible objects.
The remaining objects are evenly assigned to the two cooperative agents, then we have
$
\mathcal{I}_{\mathcal{X}} \cup \mathcal{I}_\mathcal{Y} = \mathcal{I}, \ 
\mathcal{I}_\mathcal{X} \cap \mathcal{I}_\mathcal{Y} = \mathcal{M}   
$, 
where $\mathcal{I}_{\mathcal{X}}$ and $\mathcal{I}_{\mathcal{Y}}$ denote the perceived set by the two agents.

\textbf{Perception errors. } To investigate the impact of perception errors in (\ref{eq:measurement}), different levels of
position and orientation angle errors are added to the objects in the dataset. They are set to be Gaussian distributed as
$\mathcal{N}(0,\sigma_{{p}})$ and $\mathcal{N}(0,\sigma_{{\theta}})$, respectively. 
For object detection algorithms, determining the orientation of an object is a difficult task and prone to errors. To simulate this, we added a direction noise to the orientation with a $50\%$ probability to make it face the opposite orientation.

\textbf{Localization errors. } Since the initial relative pose transformation reflects the magnitude of the pose error of the cooperating vehicles, it is set as a fixed value. 
In practice, the objects in $\mathcal{I}_\mathcal{Y}$ are translated by $3$ m in the $x$ and $y$ directions and rotated by $5\degree$ as a whole, \ie, the agents' relative position offset entirely based on accurate poses of the two vehicles. 

\subsection{Average precision of inter-agent object association}
We test the performance of benchmarks on inter-agent object association under different levels of outlier rate and perception errors (including position errors and orientation errors), the results are shown in Fig. \ref{fig:ap_matching}.

As shown in the result, $\eta$ has a more significant impact than standard deviations $\sigma_{p}$ and $\sigma_{\theta}$.
RICP exhibits a higher overall AP level than ICP, but they are both highly sensitive to $\eta$, this might be due to their use of iterative searching for the closest point in the correspondence identification step that converged to a local optimum. 
VIPS outperforms them in terms of AP, and it shows good robustness to changes in $\eta$. 
The proposed method outperforms the previous three methods in both overall precision and robustness to changes in $\eta$. 

When considering the perception errors, 
we observed that the proposed algorithm is robust to position and orientation errors, with only an overall downward shift in the AP curve at $\sigma_{{p}}=0.9$ m. For different levels of errors, the AP curve remains highly consistent with the zero error scenarios.
RICP exhibits poor robustness to position errors, 
VIPS demonstrates good robustness against position errors but is unable to deal with large orientation errors. 
This is because VIPS uses the sine difference of the heading angles of two nodes to encode edge-to-edge similarity. This makes it fragile to errors contained in the heading angle of the objects.

\section{Evaluation on cooperative perception dataset}
\label{sec:opv2v}

\subsection{Experiment setting}
\textbf{Dataset.} OPV2V \cite{Xu2022a} is a large-scale dataset that contains $73$ scenes
for V2V-based cooperative perception, including $2170$ frames for \textit{test} subset, and $549$ frames for \textit{test culver city} (\textit{tcc}). The latter is developed to narrow down the gap between the simulated and real-world traffic scenarios, which can be used to test the adaptability and portability of the proposed algorithm. 
The reasons of using OPV2V are \textit{a}) incorporation of cooperative vehicles and their locally perceived information, and \textit{b}) provision of a wide range of scenarios, including highly complex traffic scenes with numerous traffic participants.
The first row of Fig. \ref{fig:statistics on opv2v} illustrates the distributions of co-visible object rate and absolute object counts across two test sets. %

\textbf{Object detection and perception errors. }Object detection module provides inter-node measurements in (\ref{eq:measurement}). For fairness, We trained an object detection network PointPillars \cite{Lang_2019_CVPR} using the \textit{train} subset provided by OPV2V and kept it the same for all benchmarks. The performance of PointPillars on the dataset is evaluated in the second row of Fig.\ref{fig:statistics on opv2v}. 
The third subfigure of Fig. \ref{fig:statistics on opv2v} depicts the distribution of lateral ($y$) and longitudinal ($x$) position errors in the bounding boxes detected by PointPillars.
It shows that the position errors in both directions approximately follow a Gaussian distribution, with greater errors observed in the longitudinal direction, and similar error distributions are observed across both datasets.
This result supports the setting of $\sigma_p$ in SIND. The fourth subfigure of Fig. \ref{fig:statistics on opv2v} shows the distribution of angular errors in the network's perceived results. Specifically, we calculated the degree of deviation between the perceived bounding boxes and ground truth in the heading direction.
It shows that the angular errors were distributed mostly between $-20\degree$ and $20\degree$, posing a significant challenge to association algorithms. 


\begin{figure}[t]
\centering
\includegraphics[width=0.95\linewidth]{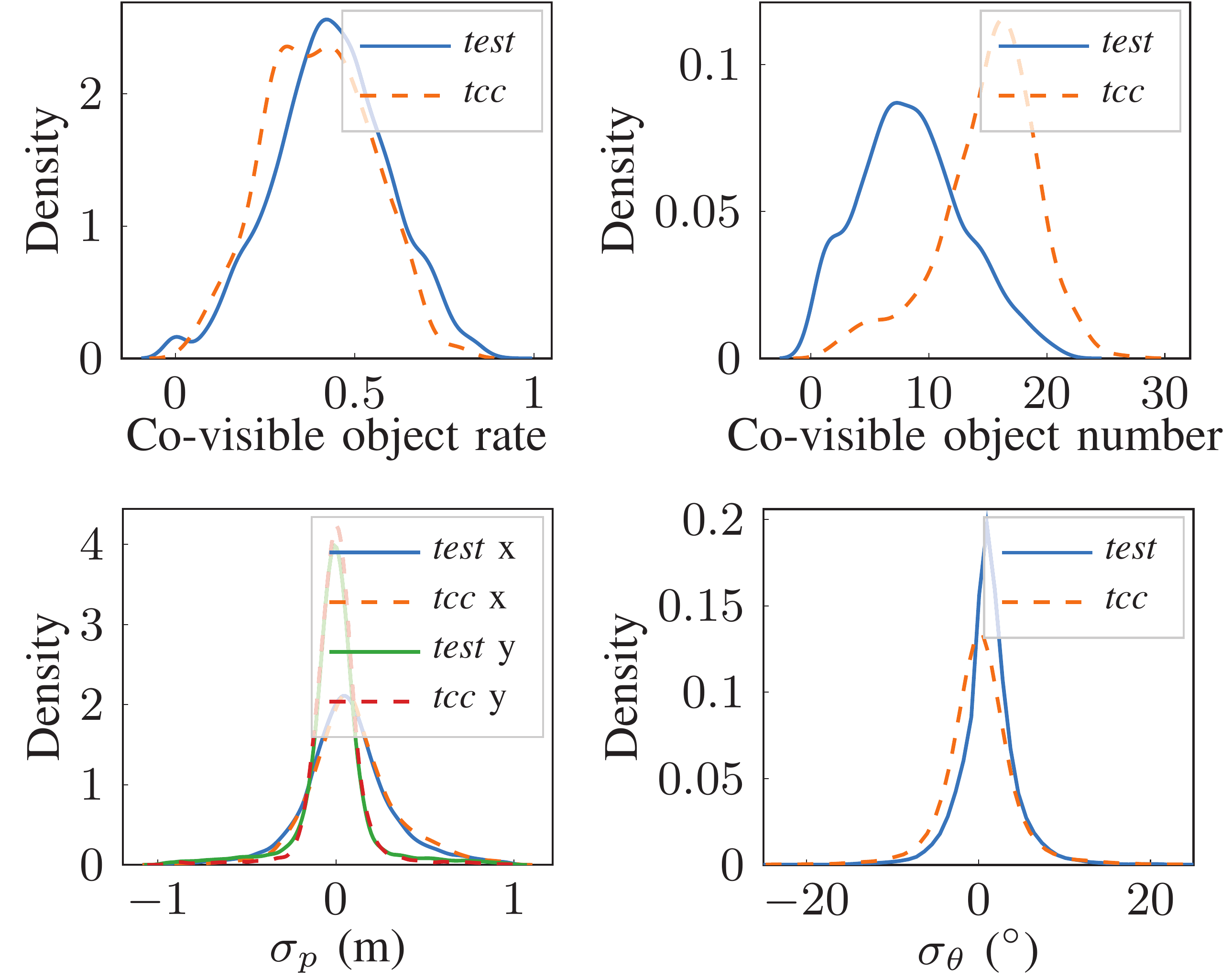}
\caption{Statistics of OPV2V dataset.}
\label{fig:statistics on opv2v}
\end{figure}

\textbf{Localization errors.}  Without loss of generality, two scenarios are considered for demonstration and comparison: the first one assumes that the participating agents have no position and orientation errors, while the second one assumes that the position and orientation errors both follow zero-mean Gaussian distribution with standard deviation $\sigma_p^L=3$ m and  $\sigma_\theta^L=5\degree$.

\subsection{Evaluation of inter-agent association performance}
\label{sec:pre}

\textbf{Precision and recall.} {The precision and recall results for the matching task on OPV2V are shown in Table. \ref{tab:ar}. }
Compared with the benchmarks, the proposed method achieves higher precision and recall and shows good robustness to the pose errors.
Note that VIPS performs significantly worse on OPV2V compared to SIND, primarily due to the complex and challenging natures of the scenarios in OPV2V, such as a larger amount of objects and perception errors. 

\textbf{Distance between correspondence pairs. }
Matching precision and recall may not be a perfect indicator of the perception performance, for example, when the two objects are close to each other, their incorrect pose estimation would not deviate significantly from the ground truth, and it would not have a major impact on the perception results. Therefore, another metric that can assess the impact of matching performance on perception is required.
We defined a metric AD (average distance) $
d=  1/N \sum_{(i,j)\in\hat{\mathcal{M}}}\text{d}\big(\bm{s}_i- \bm{s}_j\big)
$ that measures the average distance between the matched object pairs,
where $N=\text{card}( \hat{\mathcal{M}})$, and  operator  $\text{d}(\cdot)$ denotes  calculating the Euclidean distance. 
Table. \ref{tab:ar} shows the performance of the methods of $d$ on two datasets, \textit{test} and \textit{test culver city}. Notably, the $d$ values of the proposed method are quite small, which
means even for incorrectly associated object correspondences in the matching results, the distances between them are not too far away to cause fatal impacts on the estimation of the pose transformation in the back end.

\textbf{Impact of outliers. }
We present Fig. \ref{fig:cor_opv2v} depicting the distribution of average matching precision against the rate of non-co-visible objects on the \textit{test culver city} and \textit{test} datasets. The results consistently align with those obtained from the SIND dataset, as illustrated in Fig. \ref{fig:ap_matching}. This reaffirms that the proposed method exhibits remarkable resilience to outliers.

\begin{figure}[h]
\centering
\includegraphics[width=0.95\linewidth]{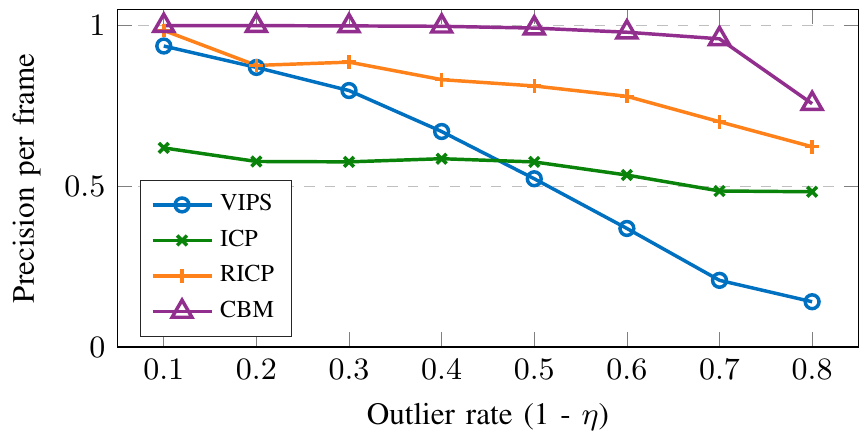}
\caption{Impact of outliers.}
\label{fig:cor_opv2v}
\end{figure}

\begin{figure}
\centering
\includegraphics[width=\linewidth]{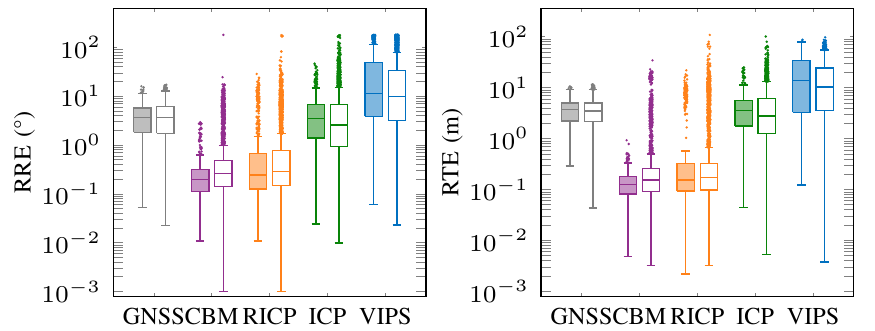}
\caption{Distribution of RRE and RTE on OPV2V.}
\label{fig:rre_rte}
\end{figure}

{
\textbf{Impact of global consensus module. }
Ablation study is implemented by substituting $\mathbf{L}_{ij}^{(2)}$ in  
(\ref{eq:solution_A}) with $\mathbf{L}_{ij}$, the result is shown in the last row of Table \ref{tab:ar}. It's observed that the absence of the global consensus module has a minor impact on precision and recall. However, it has a significant influence on the matching distance, showing an increase of approximately 35\% and 50\% on the two datasets, respectively. This indicates that global consensus is effective in distinguishing those mismatches with 
ambiguity.
}

\begin{table}[h]
\centering 
\caption{Matching performance on OPV2V with $\sigma_p^L=3 \text{ m},\sigma_\theta^L=5^\circ$. Pre.: precision, Rec.: recall, Dis.: distance. Ablation: without global consensus module. }
\small
\begin{tabular}{@{}c|c|cccccc@{}}
\toprule
\multirow{2}{*}{Dataset} & \multirow{2}{*}{Metric} & \multicolumn{5}{c}{Method} \\
& & VIPS & ICP & RICP & CBM &Ablation\\
\midrule
\multirow{3}{*}{\textit{tcc}} & Pre. (\%) & 47.8 & 46.0 & 73.6 & \textbf{99.5} &-3.1\\
& Rec. (\%) & 58.5 & 38.4 & 72.1 & \textbf{93.0}& -0.3\\
& Dis. (m) & 28.27 & 7.84 & 3.99 & \textbf{0.32} &+0.11\\
\midrule
\multirow{3}{*}{\textit{test}} & Pre. (\%) & 47.2 & 56.1 & 78.1 & \textbf{90.6} &-5.1\\
& Rec. (\%) & 57.5 & 52.2 & \textbf{80.8} & 78.0 &-0.8\\
& Dis. (m) & 22.32 & 5.53 & 2.58 & \textbf{0.50} &+0.25 \\
\bottomrule
\end{tabular}

\label{tab:ar}
\vspace{2mm}
\centering
\caption{mean Average precision  at IoU=0.7 on OPV2V  \newline under different $\sigma_p^L$.}
{\small 
\begin{tabular}{@{}c|c|cccccc@{}}
\toprule
\multirow{2}*{{Dataset}} & \multirow{2}*{{Method}} & \multicolumn{6}{c}{$\sigma_p^L$ (m)} \\
&& {0} & {0.6} & {1.2} & {1.8} & {2.4} & {3.0} \\
\midrule
\multirow{6}*{{\textit{test}}}&Single
&60.0&60.0&60.0&60.0&60.0&60.0\\
&GNSS&\textbf{80.5}&24.5&21.4&24.2&26.0&26.5\\
&ICP&44.3&37.3&32.6&28.1&26.5&26.8 \\
&VIPS&25.9&24.2&23.5&22.8&22.8&22.9\\
&RICP&71.9&\textbf{71.6}&\textbf{70.9}&\textbf{68.9}&67.4&64.3\\
&CBM &70.3&68.9&68.8&68.8&\textbf{68.9}&\textbf{68.9}\\
\midrule
\multirow{6}*{{\textit{tcc}}}&
Single&47.0&47.0&47.0&47.0&47.0&47.0\\
&GNSS&\textbf{68.4}&28.8&25.8&25.3&25.5&25.9\\
&ICP&35.8&32.2&29.3&28.3&26.9&26.8\\
&VIPS&25.7&25.8&25.6&25.7&25.8&25.8\\
&RICP&62.3&\textbf{62.3}&\textbf{62.1}&59.7&57.8&51.7\\
&CBM &62.1&61.9&61.9&\textbf{61.9}&\textbf{61.9}&\textbf{61.9} \\
\bottomrule
\end{tabular}
}   
\label{tab:ap_percep}
\end{table}


\subsection{Evaluation of transform estimation and perception}

{
The evaluation metrics for transform estimation are \textit{a}) relative rotation error $
\text{RRE} =\mathrm{arccos}\left(0.5 \cdot {\mathrm{Tr}\big(\bm{R}^T \cdot \hat{\bm{R}}\big)-0.5} \right)
$
,
and \textit{b}) relative translation error $
\text{RTE}=\|\bm{t}-\hat{\bm{t}}\|_2 
$.}
Heading errors are added to the cooperative agents for RRE estimation, and only position errors are introduced for the evaluation of RTE. The benchmark performances are compared at a fixed localization error level $\sigma_p^L=3$ m or $\sigma_\theta^L=5\degree$. The results are shown in Fig. \ref{fig:rre_rte}, where the GNSS benchmark corresponds to the results without calibration.
The proposed method has significantly reduced the median RRE to the level of $0.1\degree$ and the median RTE to the level of $0.1$ m, achieving an order of magnitude improvement over the GNSS solution and outperforming other benchmarks.  


Furthermore, the perception robustness of the methods against pose errors is evaluated. 
The evaluation metric is mean Average Perception (mAP), computed by comparing the Intersection over Union (IoU) of fused bounding boxes and the ground truth boxes.
{An IoU threshold of $0.7$ is chosen. Note that this metric is different from the average precision used in Sec. \ref{sec:pre}. 
The results are presented in Table \ref{tab:ap_percep}. Due to page limitations, only results for different $\sigma_p^L$ are shown. However, similar trends can be observed in the results for different $\sigma_{\theta}^L$.
The proposed method not only maintains a high level of performance in mAP, it is also insensitive to agent pose errors under different levels, aligning with the results in Fig. \ref{fig:rre_rte}.}

\section{Conclusions}
\label{section:conclusion}

We propose a novel object-level spatial calibration approach for connected and automated driving to address the challenges of obtaining accurate relative transformation with dynamic and random position and pose errors.
The proposed method enables robust inter-agent object association and relative pose estimation, leading to improved object-level cooperative perception. Its performance is demonstrated by extensive evaluations of the real-world dataset SIND and the cooperative perception dataset OPV2V.

\bibliographystyle{IEEEtran}
\bibliography{reference}

\end{document}